\documentclass{article}
\RequirePackage[table]{xcolor}
\usepackage{ijcai25}

\usepackage{times}
\usepackage{soul}
\usepackage{url}

\usepackage[utf8]{inputenc}
\usepackage[small]{caption}
\usepackage{graphicx}
\usepackage{amsmath}
\usepackage{amsthm}
\usepackage{booktabs}
\usepackage{algorithm}
\usepackage{algorithmic}
\usepackage[switch]{lineno}
\usepackage{CJKutf8}
\usepackage{multirow}
\usepackage{tikz}
\usepackage{amsmath}
\usepackage{enumitem}

\usepackage{tikz}

\definecolor{colorFst}{RGB}{198,224,191}
\definecolor{colorSnd}{RGB}{230,236,176}
\definecolor{colorTrd}{RGB}{255,249,187}

\newcommand{\cst}{\cellcolor{colorFst}\bf}   
\newcommand{\cnd}{\cellcolor{colorSnd}}      
\newcommand{\crd}{\cellcolor{colorTrd}}      

\definecolor{cvprblue}{rgb}{0.21,0.49,0.74}
\definecolor{ijcaired}{rgb}{1,0,0}
\usepackage[pagebackref,breaklinks,colorlinks,citecolor=cvprblue,linkcolor=ijcaired]{hyperref}

\title{Deep Learning For Point Cloud Denoising: A Survey}

\author{
Chengwei Zhang$^1$\footnote{These authors contributed equally to this work.}      \and
Xueyi Zhang$^2$\footnotemark[1]         \and
Mingrui Lao$^2$         \And
Tao Jiang$^1$           \And
Xinhao Xu$^1$           \And
Wenjie Li$^1$           \And
Fubo Zhang$^1$          \And
Longyong Chen$^1$\footnote{Correspondence to: Longyong Chen (chenly@aircas.ac.cn)}       \\
\affiliations
$^1$Aerospace Information Research Institute, Chinese Academy of Sciences\\
$^2$National University of Defense Technology\\
\emails
zhangchengwei22@mails.ucas.ac.cn,
zhangxy1998@nudt.edu.cn,
laomingrui@vip.sina.cn,\\
\{jiangtao22, xuxinhao23, liwenjie21\}@mails.ucas.ac.cn,
\{zhangfb, chenly\}@aircas.ac.cn
}

\begin{document}
\begin{CJK*}{UTF8}{gbsn} 
\maketitle
\begin{abstract}
Real-world environment-derived point clouds invariably exhibit noise across varying modalities and intensities. Hence, point cloud denoising (PCD) is essential as a preprocessing step to improve downstream task performance. Deep learning (DL)-based PCD models, known for their strong representation capabilities and flexible architectures, have surpassed traditional methods in denoising performance. To our best knowledge, despite recent advances in performance, no comprehensive survey systematically summarizes the developments of DL-based PCD. To fill the gap, this paper seeks to identify key challenges in DL-based PCD, summarizes the main contributions of existing methods, 
and proposes a taxonomy tailored to denoising tasks. To achieve this goal, we formulate PCD as a two-step process: outlier removal and surface noise restoration, encompassing most scenarios and requirements of PCD. Additionally, we compare methods in terms of similarities, differences, and respective advantages. Finally, we discuss research limitations and future directions, offering insights for further advancements in PCD.
\end{abstract}

\section{Introduction}
Point clouds serve as a fundamental 3D data format, bridging the gap between depth sensors and 3D vision algorithms. However, real-world point clouds inevitably contain noise and artifacts, which degrade the performance of downstream tasks such as point cloud classification~\cite{sun2022benchmarking}, object detection~\cite{zhang2024comprehensive}, semantic segmentation~\cite{yan2024benchmarking}, and surface reconstruction~\cite{huang2024surface}.  
Compared to traditional geometric methods, deep learning (DL)-based point cloud denoising (PCD) has gained significant attention in the research community due to its superior denoising quality and generalization capability to improve the performance of downstream tasks \cite{zhao2024triplemixer,li2024pointcvar,zhang2024pd,de2024straightpcf,mao2024denoising}.

\begin{figure}[t]
    \centering
    \includegraphics[width=\linewidth]{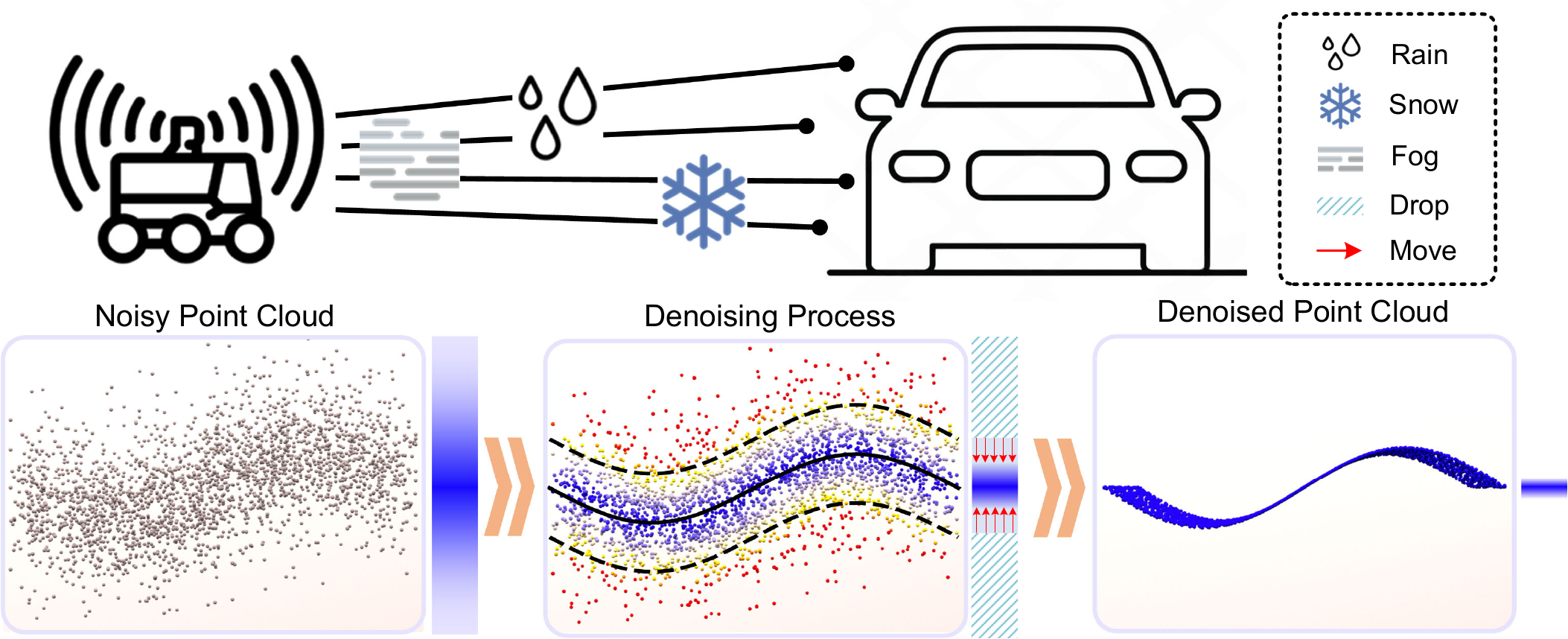}
    \vspace{-0.2in}
    \caption{Illustration of the primary causes of noisy points and the denoising process. The color bar represents the point distribution density, where darker indicate higher density. The point colors range from \textcolor{red}{red} to \textcolor{blue}{blue}, representing the distance to the underlying surface.}
    \vspace{-0.2in}
    \label{fig1}
\end{figure}

The primary causes and foundmental type of noisy points is illustrated in the upper part of Fig.~\ref{fig1}. On the one hand, the sensing imperfections and misalignment in multi-view scanning resulting points being distributed in a non-uniform manner with varying thickness and blurry geometric details, leading to \textit{surface noise} \cite{huang2024surface}. Moreover, in adverse weather conditions, active pulsed systems such as LiDAR are particularly vulnerable in scattering media such as rain, snow, and fog. Particles in these media interact with the laser beam resulting spurious returns at incorrect distances, introducing \textit{outliers} \cite{zhao2024triplemixer}. On the other hand, noise in point clouds can also be intentionally introduced with malicious intent, such as through delicately designed adversarial outliers or slight perturbations of origional surface points \cite{li2024pointcvar,sun2023critical}. 

To effectively mitigate noise, existing DL-based PCD methods can be broadly categorized into consecutive and interrelated steps: \textbf{outlier removal} and \textbf{surface noise restoration}, as illustrated in the lower part of Fig.~\ref{fig1}, which together address most scenarios and requirements of PCD. Specifically, outlier removal aims to drop points to \textit{enhance the stability of feature extraction}, and, as subsequent, surface noise restoration is designed to move points to \textit{accurately and uniformaly relocate it onto the underlying surface} while \textit{faithfully preserve fine-grained edge geometric details}.

However, existing surveys~\cite{chen2024survey,sohail2024advancing} only focus on certain aspects of DL-based PCD. \cite{chen2024survey} adopts a task-oriented perspective, primarily focusing on the graph neural network-based architectures design for surface noise restoration. \cite{sohail2024advancing} treats surface noise restoration as a subtask of point cloud enhancement, covers a limited part of surface noise restoration methods and omits the review of outlier removal methods. Moreover, their taxonomy lacks consideration of the intrinsic characteristics of denoising tasks, with insufficent discussion on the interrelations between different methods. 

To fill this gap, this survey aims to provide a comprehensive review of DL-based PCD methods.
Specifically, we consider the holistic process of outlier removal and surface noise restoration and present a generalized pipeline for DL-based PCD, covering data formats, denoising processes, and learning paradigms. Additionally, we provide a detailed overview of datasets and evaluation metrics for DL-based PCD.
Moreover, we propose a novel taxonomy for DL-based PCD, organizing existing methods in a top-down manner based on key challenges. Spercifically, this taxonomy includes two primary categories: outlier removal and surface noise restoration, with the latter further divided into edge-aware denoising and precise surface recovery. Additionally, we summarize key technical contributions from existing methods, forming a fine-grained, bottom-up classification of specific solutions. Further more, we provide a comparative analysis of their similarities, differences, and respective advantages and disadvantages.
Finally, we evaluate representative methods under a unified benchmark setting to assess their denoising performance. Based on our findings, we outline potential future research directions in DL-based PCD, including multi-task learning and improving noise generalization.

\begin{figure}[t]
\centering
\footnotesize
\resizebox*{0.5\textwidth}{!}{\begin{tikzpicture}[xscale=0.8, yscale=0.36]

\draw [thick, -] (0, 16.5) -- (0, 11); \node [right] at (-0.5, 17) {\textbf{Deep Learning on Point Cloud Denoising}};
\draw [thick, -] (0, 16) -- (0.5, 16);\node [right] at (0.5, 16) {\textbf{Outlier Removal~(Section~\textcolor{red}{\ref{sec3}})}};
\draw [thick, -] (1, 15.5) -- (1, 13);
\draw [thick, -] (1, 15) -- (1.5, 15);\node [right] at (1.5, 15) {Point-based~(Section~\textcolor{red}{\ref{sec3_1}})};
\draw [thick, -] (2, 14.5) -- (2, 14);
\draw [thick, -] (2, 14) -- (2.5, 14);\node [right] at (2.5, 14) {PointCVaR \cite{li2024pointcvar}, TripleMixer \cite{zhao2024triplemixer}};
\draw [thick, -] (1, 13) -- (1.5, 13);\node [right] at (1.5, 13) {Range-Image-based~(Section~\textcolor{red}{\ref{sec3_2}})};
\draw [thick, -] (2, 12.5) -- (2, 12);
\draw [thick, -] (2, 12) -- (2.5, 12);\node [right] at (2.5, 12) {WeatherNet \cite{heinzler2020cnn}, Slide \cite{bae2022slide}};

\draw [thick, -] (0, 11) -- (0.5, 11);\node [right] at (0.5, 11) {\textbf{Surface Noise Removal~(Section~\textcolor{red}{\ref{sec4}})}};
\draw [thick, -] (1, 10.5) -- (1, 3);
\draw [thick, -] (1, 10) -- (1.5, 10);\node [right] at (1.5, 10) {Edge-Aware Denoising~(Section~\textcolor{red}{\ref{sec4_1}})};
\draw [thick, -] (2, 9.5) -- (2, 5);
\draw [thick, -] (2, 9) -- (2.5, 9);\node [right] at (2.5, 9) {Geometry Prior};
\draw [thick, -] (3, 8.5) -- (3, 8);
\draw [thick, -] (3, 8) -- (3.5, 8);\node [right] at (3.5, 8) {GeoDualCNN~\cite{wei2021geodualcnn}, DFP~\cite{lu2020deep}};
\draw [thick, -] (2, 7) -- (2.5, 7);\node [right] at (2.5, 7) {Position-Normal Interaction};
\draw [thick, -] (3, 6.5) -- (3, 6);
\draw [thick, -] (3, 6) -- (3.5, 6);\node [right] at (3.5, 6) {PCDNF~\cite{liu2023pcdnf}, PN-Internet~\cite{yi2024pn}};
\draw [thick, -] (2, 5) -- (2.5, 5);\node [right] at (2.5, 5) {Supervision Strategy};
\draw [thick, -] (3, 4.5) -- (3, 4);
\draw [thick, -] (3, 4) -- (3.5, 4);\node [right] at (3.5, 4) {PointFilter~\cite{zhang2020pointfilter}, Re-PCD~\cite{chen2022repcd}};
\draw [thick, -] (1, 3) -- (1.5, 3);\node [right] at (1.5, 3) {Precise Surface Recovery~(Section~\textcolor{red}{\ref{sec4_2}})};
\draw [thick, -] (2, 2.5) -- (2, -9);
\draw [thick, -] (2, 2) -- (2.5, 2);\node [right] at (2.5, 2) {Denoising Path Learning};
\draw [thick, -] (3, 1.5) -- (3, 1);
\draw [thick, -] (3, 1) -- (3.5, 1);\node [right] at (3.5, 1) {Path-Net~\cite{wei2024pathnet}, StraightPCF~\cite{de2024straightpcf}};
\draw [thick, -] (2, 0) -- (2.5, 0);\node [right] at (2.5, 0) {Surface Consistency};
\draw [thick, -] (3, -0.5) -- (3, -1);
\draw [thick, -] (3, -1) -- (3.5, -1);\node [right] at (3.5, -1) {N2NM~\cite{BaoruiNoise2NoiseMapping}, SVCNet~\cite{zhao2022noise}};
\draw [thick, -] (2, -2) -- (2.5, -2);\node [right] at (2.5, -2) {Surface Field};
\draw [thick, -] (3, -2.5) -- (3, -7);
\draw [thick, -] (3, -3) -- (3.5, -3);\node [right] at (3.5, -3) {Gradient Field:};
\draw [thick, -] (4, -3.5) -- (4, -4);
\draw [thick, -] (4, -4) -- (4.5, -4);\node [right] at (4.5, -4) {Score~\cite{luo2021score}, PSR~\cite{chen2022deep}};
\draw [thick, -] (3, -5) -- (3.5, -5);\node [right] at (3.5, -5) {Diffusion Field:};
\draw [thick, -] (4, -5.5) -- (4, -6);
\draw [thick, -] (4, -6) -- (4.5, -6);\node [right] at (4.5, -6) {P2P-Bridge~\cite{vogel2025p2p}};
\draw [thick, -] (3, -7) -- (3.5, -7);\node [right] at (3.5, -7) {Neural Implicit Field:};
\draw [thick, -] (4, -7.5) -- (4, -8);
\draw [thick, -] (4, -8) -- (4.5, -8);\node [right] at (4.5, -8) {LIFPCF~\cite{wang2024learning}, N2NM~\cite{BaoruiNoise2NoiseMapping}};
\draw [thick, -] (2, -9) -- (2.5, -9);\node [right] at (2.5, -9) {Manifold Reconstruction};
\draw [thick, -] (3, -9.5) -- (3, -12);
\draw [thick, -] (3, -10) -- (3.5, -10);\node [right] at (3.5, -10) {Manifold Sampling:};
\draw [thick, -] (4, -10.5) -- (4, -11);
\draw [thick, -] (4, -11) -- (4.5, -11);\node [right] at (4.5, -11) {DMR~\cite{luo2020differentiable}, SSPCN~\cite{li2023single}};
\draw [thick, -] (3, -12) -- (3.5, -12);\node [right] at (3.5, -12) {Manifold Refinement:};
\draw [thick, -] (4, -12.5) -- (4, -13);
\draw [thick, -] (4, -13) -- (4.5, -13);\node [right] at (4.5, -13) {PD-LTS~\cite{mao2024denoising}, PD-Refiner~\cite{zhang2024pd}};

\end{tikzpicture}}
\vspace{-0.2in}
\caption{A taxonomy of representative DL-based PCD methods.}
\vspace{-0.2in}
\label{fig2}
\end{figure}
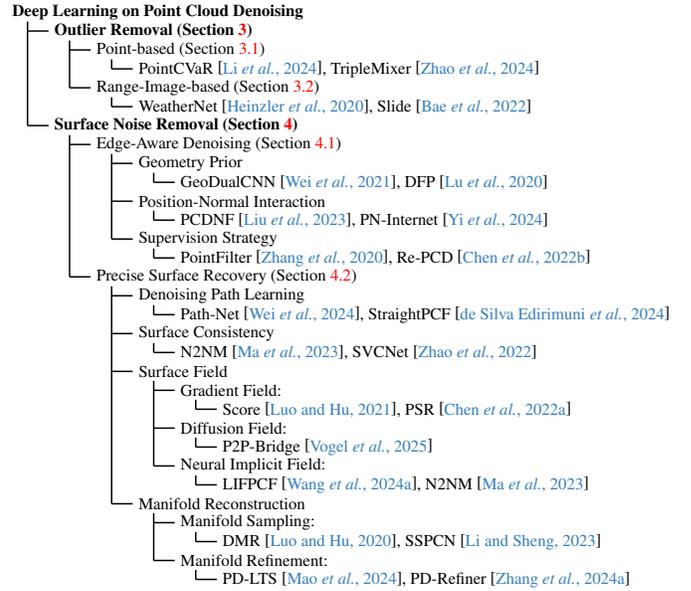

We summarize the main contributions as follows:
\begin{itemize}[leftmargin=1em]
    \item To the best of our knowledge, this is the first work that systematically and comprehensively reviews DL-based PCD methods, covering two fundamental stages: outlier removal and surface noise restoration, which encompassing most scenarios and requirements of PCD.
    \item Guided by the requirements and characteristics of PCD, we propose a taxonomy to categorize existing methods, as illustrated in Fig.~\ref{fig2}. Furthermore, we analyze the intrinsic connections between the two denoising stages and explore the underlying relationships among different approaches.
    \item We evaluate representative DL-based PCD methods in the most frequently used settings. In addition, we discuss open challenges and promising future directions to inspire further research in this field.
\end{itemize}

\begin{figure*}[ht]
    \centering
    \includegraphics[width=\linewidth]{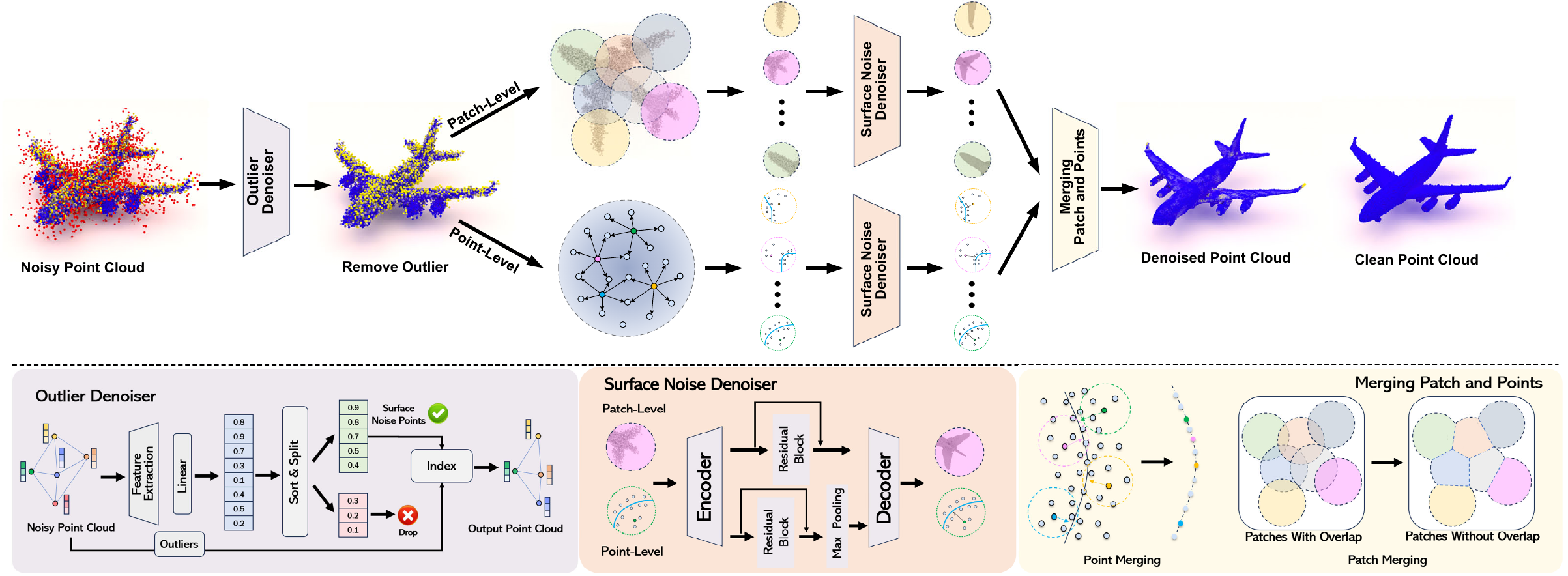}
    \vspace{-0.2in}
    \caption{An overview of DL-based PCD paradigms. The outlier denoiser can be regarded as a point-wise classification network that removes outliers based on probability estimation. The surface noise denoiser takes the outlier-denoised point cloud as input and follows either point-level or patch-level paradigms, producing denoised results either as individual points or as patch of points. Point-level denoised results are directly merged, whereas patch-level results require overlap removal before merging.}
    \label{fig3}
    \vspace{-0.1in}
\end{figure*}

\section{Background}
\subsection{Problem Formulation}
\subsubsection{Noisy Point cloud Data} 
We consider a noisy point cloud as the union of outliers and surface noise points, denoted as $P = \{p_1, p_2, \dots, p_{N+M}\}$, where $N$ and $M$ represent the number of surface noise points and outliers, respectively. Each point $p_i = [C_i, A_i]$ in the point cloud consists of two components, where $C_i$ denotes the basic 3D coordinates $(x_i, y_i, z_i)$ and $A_i$ represents the associated point attributes, e.g., color \cite{vogel2025p2p}, and normal vectors \cite{chen2023geogcn}. The denoising network can perform denoising based either solely on the spatial relationships of neighboring points or by incorporating richer contextual information from point attributes.

\subsubsection{Denoising Process}
The DL-based PCD process is illustrated in Fig.~\ref{fig3}. Our goal of interest is to recover the clean surface by denoising outlier and surface noise. Outlier removal focuses on identifying and dropping points that significantly deviate from the underlying surface. Since outlier removal relies on long-range dependencies between points, no further split is performed. This step can be viewd as an independent preprocessing stage before surface noise restoration~\cite{zhao2024triplemixer,bae2022slide}, or integrated with downstream tasks to enable end-to-end joint optimization~\cite{wang2022pointfilternet,li2024pointcvar,luo2020differentiable}.

After outlier removal, surface noise restoration aims to denoising the remaining points by moving them toward underlying surface. To enhance parallelization and accommodate the specific requirements of PCD, this step can be categorized into: Point-Level and Patch-Level. Point-Level methods constructs a smaller neighborhood graph (e.g., 128 points) for each point to independently extract features and outputs a single denoised point. After processing all points, the final denoised point cloud for the entire scene is naturally obtained \cite{wei2024pathnet}. In contrast, Patch-Level denoising selects a subset of support points from the entire point cloud using farthest point sampling (FPS), computes the K-nearest neighbors (KNN) for each support point, and split the point cloud into several overlapping, larger patches (e.g., 1024 points). The final denoised result is obtained by removing overlapping regions between patches, which is achieved through patch stitching \cite{de2023iterativepfn}.

\textbf{Discussion:} 
Point-Level and Patch-Level denoising exhibit different strengths and limitations. Specifically, the denoising results of Point-Level methods remain consistent regardless of whether the test data is complete, as the local context remains unchanged for each point. However, independently computing neighborhoods for each point introduces significant computational redundancy, which adversely affects processing efficiency. Moreover, by treating each point in isolation without considering inter-point dependencies, Point-Level denoising may lead to clustering effects, resulting in an ununiformly distributed denoised point cloud. Nevertheless, Point-Level methods can precisely focus on the denoising direction of individual points, making them well-suited for preserving fine-grained edge geometric features. As a result, most edge-aware denoising approaches belong to Point-Level methods.

Patch-Level denoising, on the other hand, is computationally more efficient and explicitly accounts for the relationships between points, leading to more uniform denoising results. Additionally, Patch-Level methods model larger-scale surface distributions, making them more effective in restoring the precise locations of underlying surfaces. However, since a single point may exhibit different denoising path in adjacent patches, inconsistencies can arise when merging patches, potentially causing artifacts such as cracks and misalignments in the final reconstructed surface.

\subsubsection{Learning Paradigm} 
The denoisers for outlier removal and surface noise restoration can be trained either separately or jointly. The outlier removal denoiser can adopt the learning paradigm of classification tasks. In supervised learning, a point-wise classification loss is typically minimized~\cite{zhao2024triplemixer}. Alternatively, unsupervised learning can be achieved by estimating reconstruction uncertainty~\cite{bae2022slide}. 

The surface noise denoiser follows a learning paradigm similar to regression tasks and is typically trained in a supervised manner by minimizing the simlarity between the denoised and ground truth (GT) point clouds~\cite{zhang2024pd} or by predicting the deviation between the estimated and GT denoising paths~\cite{de2023iterativepfn}. Additionally, unsupervised methods leverage density priors~\cite{luo2021score} or spatial consistency~\cite{BaoruiNoise2NoiseMapping} to estimate the underlying surface. 

Some approaches further exploit gradient information from backpropagated losses~\cite{li2024pointcvar,luo2020differentiable} or learn point-wise weighting schemes~\cite{wang2022pointfilternet} to jointly optimize outlier removal and surface noise restoration, enhancing overall denoising performance.

\subsection{Dataset for Point Cloud Denoising}
A large number of datasets have been collected to train and evaluate the denoisers.

For outlier removal, some methods attempt to construct physically-based models to simulate adverse environments, such as rain, snow, and fog~\cite{zhao2024triplemixer}. Others collect real-world noise points under adverse weather conditions and inject them into clean road scene datasets~\cite{bae2022slide}. Additionally, some approaches collect paired point clouds of the same scene under real-world conditions, capturing both noisy and clean versions~\cite{heinzler2020cnn}.

For surface noise restoration, there are two types of datasets: synthetic datasets and real-world datasets. Synthetic datasets typically consist of complete object-level mesh models, which are used to generate noisy point clouds with varying densities, noise modalities, and intensities~\cite{luo2021score,chen2022deep}. Real-world datasets consist of object scans~\cite{wang2016mesh} and urban street scene data~\cite{serna2014paris}, which inherently contain occlusions and more complex noise patterns.

\subsection{Evaluation Metric}
Different evaluation metrics have been proposed to evaluate the denoising methods. For outlier removal, a point-wise classification paradigm is typically adopted, leveraging commonly used classification metrics such as mean Intersection-over-Union (mIoU), precision, recall, and F1-score. These metrics are used to evaluate point-wise binary classification (determining whether a point is an outlier) as well as multi-class classification (identifying outlier types such as rain, snow, or fog). For surface noise restoration, Chamfer Distance (CD) and Point-to-Mesh Distance (P2M) are the most commonly used evaluation criteria. In particular, Chamfer Distance (CD) emphasizes the overall similarity and uniformity of the denoised point cloud, whereas Point-to-Mesh Distance (P2M) emphasizes precise surface localization.

\section{Outlier Removal}
\label{sec3}
Outliers exhibit significant sparsity differences compared to surface points, often introducing uncertainty in downstream tasks. For example, when constructing a k-nearest neighbor (KNN) graph and learning feature correlations based on differences between central and neighboring points, outliers can cause abnormally high feature activations, leading to unstable gradient propagation and parameter updates, which negatively impacts performance of downstream tasks. On the other aspect, these sparsity differences and anomalous characteristics can, in turn, serve as cues for outlier removal~\cite{bae2022slide,li2024pointcvar,zhao2024triplemixer} to enhance the stability of feature extraction. This step can directly implemented as \textit{point-based} classification on the raw point cloud, as well as, leverage sensor perspective information to project points as \textit{range images-based} pixel-level classification.

\subsection{Point-based}
\label{sec3_1}
Point-based methods offer high flexibility, making them exectuable for various types of point cloud data. Spercifically, TripleMixer~\cite{zhao2024triplemixer} treats outlier removal as an independent task and introduces the Geometry, Frequency, and Channel Mixer Layer to capture geometric spatial information, extract multi-scale frequency features, and enhance multi-channel feature representations in point clouds, providing strong interpretability for the outlier removal process. In contrast, PointFilterNet~\cite{wang2022pointfilternet} integrates outlier removal into the feature learning process and employs an outlier recognizer that assigns smaller coefficients to points farther from the underlying surface, thereby identifying and excluding them. Besides, PointCVaR~\cite{li2024pointcvar} performs gradient-based attribution analysis at the point level to estimate the risk of each point, treating high-risk points as outliers and removing them.

\subsection{Range Image-based}
\label{sec3_2}
In projected range images, outliers exhibit abrupt pixel value variations compared to their neighboring pixels, which can be effectively captured using convolutional neural networks (CNNs). To this end, WeatherNet~\cite{heinzler2020cnn} presents the first CNN-based approach to analyze and filter out adverse weather effects in point cloud data. Besides, Slide~\cite{bae2022slide} leverages the structural characteristics of noisy points, particularly their low spatial correlation with neighboring points, to develop a novel self-supervised outlier removal framework. Further more, LiSnowNet~\cite{yu2022lisnownet} integrates Fast Fourier Transform (FFT) with CNNs to perform unsupervised denoising.

\section{Surface Noise Removal}
\label{sec4}
Outlier removal can only drop noisy points without adjusting their positions, leaving the remaining point cloud around the underlying surface unevenly distributed and with a certain thickness. Thus, surface noise restoration is employed to adjust the positions of the rest noisy points, enabling them to accuratly and uniformly converge onto the underlying surface. In this field, researchers primarily focus on accurately estimating the surface location from the point cloud while preserving sharp geometric edges during denoising. Based on these objectives, we categorize surface noise restoration methods into two main groups: \textit{Edge-Aware Denoising} and \textit{Precise Surface Recovery}.

\subsection{Edge-Aware Denoising}
\label{sec4_1}
The edge regions of point clouds contribute more to the description and identification of structures \cite{wu_2023_attention}. To this end, edge-aware denoising methods enhance the preservation of edge features by leveraging surface geometric information from three key aspects: \textit{Geometry Prior}, \textit{Position-Normal Interaction}, and \textit{Supervision Strategy}.

\subsubsection{Geometry Prior}
\label{sec4_1_1}
The edge regions, where two surfaces meet, provide geometric prior information to guide edge feature restoration. GeoDualCNN~\cite{wei2021geodualcnn} uses a dual convolutional network with geometric support to compute a homogeneous neighborhood for each point. DFP~\cite{lu2020deep} uses a classifier to distinguish between anisotropic feature points and isotropic non-feature points. These points are used to construct separate height maps for training a normal estimation network, which updates the point positions. Although effective, these methods rely on point-wise priors, which are often oversimplified and less generalizable across different models.

\subsubsection{Position-Normal Interaction}
\label{sec4_1_2}
Normals are an intuitive way to describe local geometric properties, and the interaction between the denoising process and normal estimation will help denoiser learn richer geometric representations. 
For example, GeoGCN~\cite{chen2023geogcn} introduces a cascaded single-stream network. The first stage aims for coarse denoising and normal estimation while the second stage responsible for fine-grained denoising.
Further more, dual-stream networks can enable mutual learning between point cloud denoising and normal estimation. PCDNF~\cite{liu2023pcdnf} first estimates normals from the noisy point cloud and then develops a point-normal feature interaction network within a joint task paradigm to simultaneously learn point cloud denoising and normal filtering. Besides, PN-Internet~\cite{yi2024pn} designs two interactive graph convolutional networks (GCNs) to exploit the geometric dependencies between point positions and normals.

\subsubsection{Supervision Strategy} 
\label{sec4_1_3}
Incorporating additional loss terms or weighting strategies based on geometric structures during loss computation can promote the network focus more on edge region. In the early research, PointFilter~\cite{zhang2020pointfilter} computes the normal of the ground truth point nearest to the filtered point and uses its similarity to the patch center normal as a loss term. More directly, CL~\cite{de2023contrastive} employs a single-stream feature encoder to generate latent representations of patches based on patch similarity and a regressor to infer both point normals and displacements simultaneously. The predicted positions and normals are jointly optimized to leverage geometric normal information.

During the loss aggragtion process, some methods attempt to assign fixed point-wise weights or perform adaptive edge-aware weighting. For instance, Re-PCD~\cite{chen2022repcd} assigns fixed weights to each point based on surface curvature variations, reflecting their varying contributions to the loss function. In contrast, PD-Refiner~\cite{zhang2024pd} learns edge information from the ground truth point cloud by modeling geometric variations through variance in feature attention maps, without requiring explicit normal supervision during training. Further more, some methods explore transferring edge information from the image domain to the point cloud domain. 3DMambaIPF~\cite{zhou20243dmambaipf} introduces a fast differentiable rendering loss, which enhances the visual realism of denoised geometric structures by computing the discrepancy between rendered images of the denoised and ground truth point clouds.

\subsection{Precise Surface Recovery}
\label{sec4_2}
Accurate position of underlying surface is needed to guide noisy points convergence uniformly onto it. To achieve precise surface recovery, researchers have proposed several effective strategies, such as \textit{Denoising Path Learning}, \textit{Surface Consistency}, \textit{Surface Field}, and \textit{Manifold Reconstruction}.

\subsubsection{Denoising Path Learning}
\label{sec4_2_1}
The process of moving noisy points toward the underlying surface generally cannot be completed in a single step but requires multiple iterations to achieve stable convergence. Several methods explore the contextual dependencies in denoising path to improve geometric consistency during the iterative process. Re-PCD~\cite{chen2022repcd} introduces a recurrent neural network (RNN) to perform sequence modeling across denoising stages, ensuring that fine-grained geometric details are retained or recovered during iterations. Recently, $C^2AENet$~\cite{chen2024progressive} enhances interaction across multiple denoising stages by introducing cross-stage cross-coder connections in a cascaded denoising network. 

Besides, adaptive path length and straightened denoising paths facilitate faster convergence to the underlying surface. On the one hand, Path-Net~\cite{wei2024pathnet} employs reinforcement learning to design dynamic deep denoising paths that adapt to varying noise intensities. On the other hand, StraightPCF~\cite{de2024straightpcf} focuses on improving the straightness of denoising path. 

Furthermore, segmenting the denoising process into multiple intermediate stages with prograssive objectives can alleviate the learning burden of each stage. IterativePFN~\cite{de2023iterativepfn} introduces a stacked IterationModule to model the iterative denoising process in an end-to-end manner. Based on that, they progressively adds noise to the ground truth (GT) point cloud, decreasing the noise level with each iteration, allowing each IterationModule to focus on different denoising stages and learn shorter denoising path. Notably, adaptive loss strategies have also been widely adopted in subsequent works~\cite{mao2024denoising,chen2024progressive,zhou20243dmambaipf}. 

Different from previous supervised methods, learning denoising path under unsupervised manner have also been explored. As the basic, Noise4Denoise~\cite{wang2024noise4denoise} proposes an unsupervised framework where the existing noisy point cloud serves as the optimization target. It generates an input with twice the noise level and learns the denoising path by scaling the position displacement vectors by a factor of two to obtain the final output. Further more, SITF~\cite{su2025sitf} extends this approach into an iterative framework, refining the denoising path through multiple iterations.

\subsubsection{Surface Consistency}
\label{sec4_2_2}
Point clouds obtained from multiple observations of the same surface exhibit different noise signal while implicitly encoding consistent underlying surface information. Specifically, these multiple observations can be acquired at different time instances, necessitating the consideration of both temporal variability and invariance. 

On the one hand, LiDAR systems can capture multiple noisy point cloud instances of the static surface. Based on this principle, N2NM~\cite{zhou2024fast} leverages noise-to-noise mapping to learn the temporally invariant geometric consistency from multiple observations of static objects to denoising without requiring clean point clouds. On the other hand, other methods explore consistency in dynamic scenes, where sensor positions and object locations change over time. For example, VideoPD~\cite{hu2022dynamic} exploits temporal correspondences by estimating gradient fields \cite{luo2021score}, where temporally corresponding patches are identified by leveraging rigid motion principles from classical mechanics. Besides, 4DenoiseNet~\cite{seppanen20224denoisenet} adopts a scene flow estimation paradigm, proposing a temporal KNN convolution to perform point cloud denoising by leveraging cross-frame surface consistency and noise inconsistency in consecutive point cloud sequences.

Moreover, multiple observations can be synthesized manually by introducing noise variations. To realize this, SVCNet~\cite{zhao2022noise} generates noise variations at the feature level and employs a cross-attention mechanism to capture commonalities with the original features. Simlarily, CL~\cite{de2023contrastive} introduces noise perturbations in input patches as an augmentation strategy to simulate multiple observations of the underlying surface. By incorporating contrastive learning, it pulls together representations of patches sharing the same underlying clean structure.

\subsubsection{Surface Field}
\label{sec4_2_3}
The distribution of surfaces noise points can be modeled by various fields, facilitating point displacement for denoising.

\textbf{Gradient Field:} 
Surface points exhibit a distributional property where density is higher near the surface and lower farther away. Based on this property, Score~\cite{luo2021score} models the point cloud distribution as a gradient field and increases the log-likelihood of each point via gradient ascent, iteratively updating its position. Further, PSR~\cite{chen2022deep} introduced a gradient-based point set resampling approach that continuously models the probability density distribution of noisy point clouds and resamples this distribution to obtain a denoised point cloud. Besides, MAG~\cite{zhao2023point} introduced a momentum term in the gradient update process to mitigate fluctuations in the gradient field. In particular, GPCD++~\cite{xu2022gradient} identified that gradient-based methods often assume point independence, leading to uneven distributions such as holes and clustering. To address this, they proposed an ultra-lightweight network at the backbone’s endpoint to model local point interactions.

\textbf{Diffusion Field:}  
Although diffusion models are widely applied in shape generation, in surface noise restoration, researchers focus on enhancing their accuracy and stability in surface recovery. Recently, P2P-Bridge~\cite{vogel2025p2p} approaches the denoising task by formulating it as a Schrödinger bridge problem, solving it by training a network to determine an optimal transport plan between the noisy and corresponding clean point clouds. Further more, LHD~\cite{xu2024point_LHD} proposes a learnable heat diffusion framework. Unlike previous diffusion models with a fixed prior, its adaptive conditional prior selectively preserves geometric features of the point cloud by minimizing a refined variational lower bound, guiding points to evolve toward the underlying surface during the reverse process.

\textbf{Neural Implicit Field:} 
Gradient and diffusion fields rely on discrete local point neighborhoods, limiting their representational continuity. In contrast, neural implicit fields provide a more continuous surface representation. Specifically, implicit functions, i.e., signed distance functions (SDFs) and unsigned distance functions (UDFs), have been widely used for point cloud denoising, as they naturally indicate the distance from any location in 3D space to the nearest surface. Based on that, the denoising process treats noisy points as query points, utilizing the learned implicit function to predict distances and adjusting their positions via gradient-based updates.

On the one hand, LIFPCF~\cite{wang2024learning} proposes a data-driven method to learn SDFs. Similar to point-level denoising, it predicts a 7D vector for each local patch, representing the SDF of a support point. The distance can be directly obtained from the first element of the vector, while the movement direction is determined by computing the gradient descent from the last six elements (i.e., six surrounding SDFs). On the other hand, N2NM~\cite{zhou2024fast} introduces an overfitting-based approach that learns SDFs from noisy point clouds via noise-to-noise mapping for statistical reasoning. Combining the advantages, LocalN2NM~\cite{chen2024inferring} intergrates the efficiency and generalization capabilities of data-driven and overfitting-based methods. Specifically, by fine-tuning data-driven priors to learn SDFs, it achieves high-precision and efficient SDFs inference.

In summary, implicit functions are well-suited for handling sparse and high-noise-intensity point clouds due to the following reasons:  
1) Implicit functions provide a continuous surface representation, allowing for enhanced point cloud continuity by generating and denoising additional query points~\cite{ouasfi2024mixing}.  
2) Implicit functions enable larger positional displacements, as they estimate spatial distances over a broader range rather than relying solely on local neighborhoods. This property allows them to better handle high-intensity and complex noise modalities in point clouds~\cite{zhou2024fast,chen2024inferring}.

\subsubsection{Manifold Reconstruction}  
\label{sec4_2_4}
The underlying manifold of surface noise points can be encoded into point-wise features through the feature extraction process. After that, the manifold reconstruction can be performed to denoise the point cloud by applying sampling and refinement processes to the manifold representation.

\textbf{Manifold Sampling:} 
The manifold downsampling is the primary denoising step with certain robustness to outliers, while the manifold upsampling is responsible for the restoration of continuous surfaces.
Based on this, DMR~\cite{luo2020differentiable} introduces differentiable pooling, scoring each point based on noise intensity and selecting low-noise points. It then restores continue and low noise manifold by point split along the surface. Further more, SSPCN~\cite{li2023single} extends DMR by incorporating feature compensation during intermediate stages. 
Additional, such up and down sampling steps can build a pyramid denoiser.
On the one hand，Random~\cite{wang2023random} only employs down-sampling for feature aggregation based on random selection, integrating features from dense and sparse points to embed rich geometric information from dense regions into sparse point representations. On the other hand, PyramidPCD~\cite{liu2024pyramidpcd} proposes a U-Net-like structure for point clouds, enabling multi-scale geometric perception through a pyramid-based learning framework.

\textbf{Manifold Refinement:}
Similar to the modification of point position, the representations can also be refined.
Such refinement can only focus on noise component. PD-Flow~\cite{mao2022pd} employs normalized flows to map 3D space into a manifold space, explicitly disentangling noise components to extract clean latent encodings and inverse this process to recover denoised points. Further more, PD-LTS~\cite{mao2024denoising} incorporates rich geometric structural features within an invertible neural network to guide the noise decoupling process. Also, some methods attenpts to perform refinement among all channels. With the help of knowledge distillation, FCNet~\cite{wang2023fcnet} learns a clean manifold representation from noise-free point clouds and transfers this knowledge into a student network trained on noisy point clouds, promoting noise-free manifold learning. Besides, PD-Refiner~\cite{zhang2024pd} proposes a underlying surface inheritance and refinement paradigm, where the estimated manifold from a preceding denoiser is inherited and strengthened by incorporating multi-scale neighborhood information. Considering previous methods adopt discrete integer step sizes in graph convolution, GD-GCN~\cite{xu2024point_GD_GCN} introduces micro-step precision to realize stable manifold correction from a dynamical system perspective.

\begin{table*}[t]
\centering
\resizebox{7in}{!}{\begin{tabular}{cccccccccccccc}
\toprule[.3ex]
\multicolumn{2}{c}{\textbf{\#Points}}                     & \multicolumn{6}{c}{\textbf{10K Points}}                                                                           & \multicolumn{6}{c}{\textbf{50K Points}}                                                                                                                                                                                               \\\midrule[.2ex]
\multicolumn{2}{c}{\textbf{Noise Level}} & \multicolumn{2}{c}{\textbf{1\% Noise}} & \multicolumn{2}{c}{\textbf{2\% Noise}} & \multicolumn{2}{c}{\textbf{3\% Noise}} & \multicolumn{2}{c}{\textbf{1\% Noise}} & \multicolumn{2}{c}{\textbf{2\% Noise}} & \multicolumn{2}{c}{\textbf{3\% Noise}} \\\cmidrule[.2ex](r){1-4}\cmidrule[.2ex](lr){5-6}\cmidrule[.2ex](lr){7-8}\cmidrule[.2ex](lr){9-10}\cmidrule[.2ex](lr){11-12}\cmidrule[.2ex](l){13-14}                                
\textbf{Method}              & \textbf{Venue}              & \textbf{\textit{CD}\textdownarrow}               & \textbf{\textit{P2M}\textdownarrow}              & \textbf{\textit{CD}\textdownarrow}               & \textbf{\textit{P2M}\textdownarrow}              & \textbf{\textit{CD}\textdownarrow}               & \textbf{\textit{P2M}\textdownarrow}              & \textbf{\textit{CD}\textdownarrow}               & \textbf{\textit{P2M}\textdownarrow}              & \textbf{\textit{CD}\textdownarrow}               & \textbf{\textit{P2M}\textdownarrow}              & \textbf{\textit{CD}\textdownarrow}               & \textbf{\textit{P2M}\textdownarrow}      \\
    \midrule[.2ex]
    \bf DMR \cite{luo2020differentiable}   & \bf ACM MM 20 & 44.82 & 17.22 & 49.82 & 21.15 & 58.92 & 28.46 & 11.62 & 4.69  & 15.66 & 8.00  & 24.32 & 15.28 \\
    \bf PointFilter \cite{zhang2020pointfilter}   & \bf TVCG 20 & 28.67 & 7.54 & 39.79 & 13.06 & 49.44 & 20.76 & 6.92 & 1.15  & 11.16 & 4.06  & 15.42 & 7.23 \\
    \bf Score \cite{luo2021score} & \bf ICCV 21 & 25.21 & 4.63  & 36.86 & 10.74 & 47.08 & 19.42 & 7.16  & 1.50  & 12.88 & 5.66  & 19.28 & 10.41 \\
    \bf PD-Flow \cite{mao2022pd} & \bf ECCV 22 & 21.26 & 3.81  & 32.46 & 10.10 & 44.47 & 19.99 & 6.51  & 1.64  & 11.73 & 5.81  & 19.14 & 12.10 \\
    \bf PSR \cite{chen2022deep}   & \bf TPAMI 22 & 23.53 & 3.06  & 33.50 & 7.34  & 40.75 & 12.42 & 6.49  & 0.76  & 9.97  & 2.96  & \crd 13.44 & \cst 5.31 \\
    \bf N2NM \cite{BaoruiNoise2NoiseMapping} & \bf ICML 23 & \cst 10.60 & 2.41  & 29.25 & 10.10  & 42.21 & 18.47 & \cst 3.77 & 1.55  & 10.29 & 4.84  & 16.54 & 9.72 \\
    \bf IPFN \cite{de2023iterativepfn} & \bf CVPR 23 & 20.56 & \crd 2.18 & 30.43 & \crd 5.55  & 42.41 & 13.76 & 6.05  & \crd 0.59  & 8.03  & \cnd 1.82  & 19.71 & 10.12 \\
    \bf PathNet \cite{wei2024pathnet} & \bf TPAMI 24 & 26.72 & 5.84 & 39.73 & 12.99 & 45.24 & 24.04 & 7.16  & 1.24  & 11.40 & 4.10  & 18.75 & 9.52 \\
    \bf PD-Refiner \cite{zhang2024pd}& \bf ACM MM 24 & \cnd 17.54 & \cst 1.66 & \cnd 24.44 & \cst 4.49 & \cst 30.77 & \cst 9.13 & \cnd 4.66  & \cst 0.45 & \cnd 6.53 & \cst 1.64 & \cst 12.28 & \cnd 5.75 \\
    \bf StraightPCF \cite{de2024straightpcf}& \bf CVPR 24 & 18.70 & 2.39 & \crd 26.44 & 6.04 & \cnd 32.87 & \cnd 11.26 & 5.62  & 1.11 & \crd 7.65 & \crd 2.66 & \cnd 13.07 & \crd 6.48 \\
    \bf PD-LTS \cite{mao2024denoising}& \bf CVPR 24 & \crd 17.81 & \cnd 1.82 & \cst 24.41 & \cnd 4.70 & \crd 34.33 & \crd 11.98 & \crd 4.70  & \cnd 0.54 & \cst 6.46 & \cnd 1.82 & 18.52 & 10.67 \\
    \bf P2P-Bridge \cite{vogel2025p2p}& \bf ECCV 24 & 22.85 & 3.94 & 32.03 & 8.10 & 39.89 & 14.18 & 5.86  & 0.85 & 9.02 & 3.25 & 15.56 & 8.45 \\
    \bottomrule[.3ex]
    \end{tabular}}%
\caption{Denoising comparison with SOTA methods in PU-Net dataset under Gaussian noise. CD is multiplied by $10^5$ and P2M is multiplied with $10^5$. Data in bold and underline represent the best and second best result among all methods.}
\vspace{-0.2in}
\label{table1}%
\end{table*}%

\subsection{Performance Comparation}
Following the settings of previous work \cite{luo2021score}, we select representative and milestone methods for comparison, as shown in Table \ref{table1}. Patch-level methods \cite{de2024straightpcf,mao2024denoising,zhang2024pd} generally exhibit better denoising performance compared to point-level methods \cite{wei2024pathnet,zhang2020pointfilter}. DMR \cite{luo2020differentiable} demonstrates certain outlier removal capability, though it may sacrifice some performance in surface noise restoration. Normolization flow-based methods \cite{mao2024denoising,mao2022pd} not perform well under high noise levels, likely due to their limited generalizability. IPFN \cite{de2023iterativepfn} provides a robust backbone network for denoising, which has been followed up by many subsequent works \cite{de2024straightpcf,mao2024denoising,zhang2024pd,zhou20243dmambaipf}. Gradient field and diffusion field-based methods \cite{vogel2025p2p,luo2021score,chen2022deep} show some potential in denoising high-noise point clouds. Notably, N2NM \cite{BaoruiNoise2NoiseMapping} is primarily based on overfitting, which results in lower Chamfer Distance (CD) scores under low-noise.

\section{Conclusions and Future Directions}
This paper provides a comprehensive overview of DL-based PCD. We consider DL-based PCD as a joint process of outlier removal and surface noise restoration, and providing the corresponding general processing paradigm. We introduce widely used datasets and standard evaluation metrics. Furthermore, we propose a taxonomy that aligns with the characteristics of the PCD task, introduce representative methods, and analyze their interrelationships. Despite the notable achievements of PCD models, several challenges remain, highlighting promising directions for future research.

\subsection{Multi-Task Learning}
Most existing PCD methods rely solely on clean point clouds as reference. Future research can explore integrating denoising with both low-level and high-level tasks for more comprehensive frameworks.

\subsubsection{Low-Level Multi-Task Learning}
PCD shares complementary objectives with tasks such as surface reconstruction, normal estimation, and upsampling. Both denoising and surface reconstruction aim to learn the underlying surface from noisy point clouds~\cite{BaoruiNoise2NoiseMapping,huang2024surface}. Implicit neural field-based methods enable noise-robust, continuous representations, which aid learning in high-noise scenarios. Additionaly, normal estimation and upsampling refine surface attributes, and denoising can improve their performance when integrated these attributes~\cite{chen2022deep,de2023contrastive}.

\subsubsection{High-Level Multi-Task Learning}
High-level tasks such as classification, detection, and segmentation involve part-level structural learning. Current denoising methods primarily focus on local features, whereas part-level representations capture object structures (e.g., legs of a table or wings of an airplane). Integrating high-level tasks into denoising models can propagate part-level information through gradients or knowledge distillation, enhancing the model’s ability to capture larger-scale structures.

\subsection{Generalization}
Current methods typically generate training samples by adding Gaussian noise within a predefined low intensity range. However, these synthetic noise distributions have limited variability and intensity coverage, resulting in poor generalization to real-world scenarios.

\subsubsection{Noise Modality Shift}
During inferencing stage, there are different noise distributions and higher noise intensities, leading to performance degradation on unseen noise modalities~\cite{chen2022deep} and real-world millimeter-wave or synthetic aperture radar (SAR) point clouds, due to inconsistencies between the learned Gaussian-distributed density variations and the actual noise distributions encountered during inferencing.

\subsubsection{Beyond Fully Supervised Learning}
Large-scale real-world point cloud data without GT surface is increasingly captured using LiDAR, millimeter-wave radar, or SAR. Current supervised SOTA denoising methods fail to leverage these vast amounts of unlabeled data, making it difficult to learn noise characteristics from real-world point clouds and improve generalization. Unsupervised learning can address this limitation by leveraging large-scale unlabeled real-world point cloud data to construct self-supervised or generative learning frameworks, better modeling and adapting to real-world noise characteristics.

\clearpage
{
    \small
    \bibliographystyle{named}
    \bibliography{ijcai25}
}
\end{CJK*} 
\end{document}